\RequirePackage{amsmath}
\documentclass[runningheads]{llncs}
\usepackage{graphicx}
\usepackage[mode=buildnew]{standalone}
\usepackage{amssymb}
\usepackage{subcaption}
\begin{document}
\title{SwInception - Local Attention Meets Convolutions}
%
%
\author{David Hagerman\inst{1}\orcidID{0000-0002-5193-3789} \and
Roman Naeem\inst{1}\orcidID{0009-0001-2625-125X} \and
Jakob Lindqvist\inst{1}\orcidID{0000-0003-2790-8775} \and
Carl Lindström\inst{1,2}\orcidID{0009-0006-3563-8946} \and 
Fredrik Kahl\inst{1}\orcidID{0000-0001-9835-3020} \and
Lennart Svensson\inst{1}\orcidID{0000-0003-0206-9186}}
\authorrunning{D. Hagerman et al.}
%
\institute{Chalmers University of technology, 412 96 Gothenburg, Sweden \and
Zenseact, Lindholmspiren 2, 417 56 Gothenburg, Sweden}
\maketitle              
\begin{abstract}
Sparse vision transformers have gained popularity as efficient encoders for medical volumetric segmentation, with Swin emerging as a prominent choice. Swin uses local attention to reduce complexity and yields excellent performance for many tasks but still tends to overfit on small datasets. To mitigate this weakness, we propose a novel architecture that further enhances Swin's inductive bias by introducing Inception blocks in the feed-forward layers. The introduction of these multi-branch convolutions enables more direct reasoning over local, multi-scale features within the transformer block. We have also modified the decoder layers in order to capture finer details using fewer parameters. We demonstrate a performance improvement on eleven different medical datasets through extensive experimentation. We specifically showcase advancements over the previous state-of-the-art backbones on benchmark challenges like the Medical Segmentation Decathlon and Beyond the Cranial Vault. By showing that the existing inductive bias in Swin can be further improved, our work presents a promising avenue for enhancing the capabilities of sparse vision transformers for both medical and natural image segmentation tasks. Code and pre-trained weights can be accessed at https://github.com/Eiphodos/SwInception.

\keywords{Vision transformers \and Medical images \and Convolutional Neural Networks.}
\end{abstract}
\section{Introduction}
\label{sec:intro}

Vision Transformers (ViTs) \cite{vit} have emerged as a promising alternative to convolutional neural networks (CNNs) for many vision tasks. It is based on the transformer architecture and uses attention \cite{vaswani_attention_2017} mechanisms to capture long-range dependencies and global context in images. A ViT model can be pre-trained in a self-supervised fashion using using masked image modeling \cite{he_masked_2021}, but the architecture suffers from long convergence times, large data requirements, and high computational complexity. 

Various attempts have been made to mitigate the limitations of ViT, including the introduction of sparse attention mechanisms such as in the Swin Transformer (Swin) \cite{swin}. The local attention used in Swin reduces the computational complexity and yields an inductive bias by only attending to nearby features. Swin also utilizes a shifted window scheme to enable cross-windows connections over the local windows.
Compared to ViT, sparse transformers generally perform better on moderate-sized datasets. Nonetheless, the inherent inductive bias is limited to the attention windows, which in domains like medical volumetric segmentation are often restricted to a small size due to the high GPU memory requirements.


In this paper, we introduce an encoder architecture dubbed SwInception that attempts to alleviate these weaknesses. SwInception is a hybrid model that combines the strengths of transformers and convolution layers in a multi-branch approach.
The use of convolutions provides a stronger local inductive bias that leads to faster convergence, more accurate predictions, and reduced data requirements. Additionally, incorporating branches with receptive fields of multiple scales enhances the capacity of transformer blocks to process features of different sizes effectively. We also modify the decoder in the previous state-of-the-art model to more efficiently utilize the feature vectors from the encoder. Extensive experiments conducted on a range of medical datasets representing various modalities demonstrate a substantial advancement over the preceding state-of-the-art methodologies.

The main contributions can be summarized as follows:
 \begin{enumerate}
\item We show that the inductive bias of the sparse transformer Swin can be improved further by introducing convolutions in the feed-forward blocks.
\item We present a novel encoder architecture based on these findings named SwInception that has an improved inductive bias, a larger receptive field, and sub-layer multi-scale features.
\item We improve upon an existing decoder for medical volumetric segmentation, resulting in a model with fewer parameters and better performance.
\item Experimental results demonstrate improved performance compared to the previous state-of-the-art on two publicly available benchmarks, the Medical Segmentation Decathlon (MSD) and Beyond the Cranial Vault (BTCV).
\end{enumerate}

\section{Related Work}
\label{sec:related_work}


\noindent\textbf{Efficient Vision Transformers.} 
The ViT suffers from quadratic complexity and there are various methods to reduce it to linear complexity to enable larger inputs. Some works \cite{choromanski_rethinking_2020,kitaev_reformer_2020} approximate the attention operation with a less computationally expensive one. While, others \cite{xie_segformer_2021,wang_linformer_2020} approximate the softmax operation inside the attention operator by replacing the key and value matrices with a low-rank approximation.

\noindent\textbf{Sparse Vision Transformers.}
The predominant approach, also employed in our paper, involves reducing complexity through sparse attention.
Extensive efforts \cite{swin,ho_axial_2019,dong_cswin_2022,yang_focal_2021} have been dedicated to linearizing the attention operation by incorporating sparseness, with each work employing different selection methods for tokens to attend to. Among these methods, sparse local attention stands out for vision tasks due to its inductive local bias. However, enlarging their window size is costly and the receptive field can often be relatively small. Among sparse transformers, Swin is widely adopted, primarily for its shifted window strategy that enhances the receptive field. While state-of-the-art vision models pre-trained on massive datasets like JFT \cite{zhai_scaling_2022} often use some kind of ViT variant as encoders, Swin-based models are frequently preferred in domains with smaller datasets, such as medicine.

\noindent\textbf{Transformer-Convolution hybrid models.} Considering the complementary properties of CNNs and transformers, it is natural to want to combine the two architectures. Several studies have investigated using depth-wise convolutions in the feed-forward layer \cite{yuan_incorporating_2021,guo_cmt_2022,xie_segformer_2021} within a ViT. Depth-wise convolutions have also both been used to project attention matrices \cite{wu_cvt_2021} and to directly produce attention weights \cite{meng_segnext}. Depth-wise convolutions are cheap but lack the ability to use information from different channels, limiting their expressiveness.
In \cite{zhang_vitaev2_2023,si_inception_2022}, convolutions are explored using a separate parallel branch to the multi-head self-attention, similar to Inception \cite{szegedy_going_2014}. However, in both works, the interlayer fusion between the CNN and transformer features is applied to downsampled features due to the full attention.

\noindent\textbf{Medical volumetric segmentation.} Until recently, the state-of-the-art in medical volumetric segmentation were models identified using network search \cite{He_2021_CVPR,isensee_nnu-net_2021}. While vision transformers have become state-of-the-art for image classification, medical volumetric segmentation cannot utilize full attention as easily. The extra dimension introduces an order of magnitude more input tokens, which is problematic for models with quadratic computational complexity. The complexity can be alleviated by using smaller models and larger patch embeddings \cite{hatamizadeh_unetr_2021}, but most medical segmentation tasks require voxel-level precision where large patches are counter-productive. Newer models, therefore, use transformers with some way of enforcing locality \cite{rahman_medical_2023,huang_missformer_2023} and most of them \cite{tang_self-supervised_2022,zhou_nnformer_2022,liu_phtrans_2022,lin_contrans_2022} use a Swin-based encoder due to its lower computational complexity. Recent research \cite{liu_phtrans_2022,lin_contrans_2022} has examined the combination of Swin and convolutions; however, this exploration has been limited to parallel integration with transformer blocks, which prevents the localized enhancement of features within the blocks. The top ranking architecture on MSD, the Universal Model \cite{liu2023clip}, utilizes SwinUNETR \cite{tang_self-supervised_2022} as the base segmentation model and improves performance by significantly increasing the size of the pre-training dataset through the use of CLIP embeddings. 


\section{Methodology}
\label{sec:methodology}
 


In this section, we introduce the SwInception architecture, which uses a UNet structure with multi-scale features. An overview of the proposed architecture is shown in Fig.~\ref{fig:swinception_arch}, and additional details about the SwInception encoder and decoder are given below. 

\begin{figure*}[ht]
\centering
\includegraphics[width=.90\linewidth]{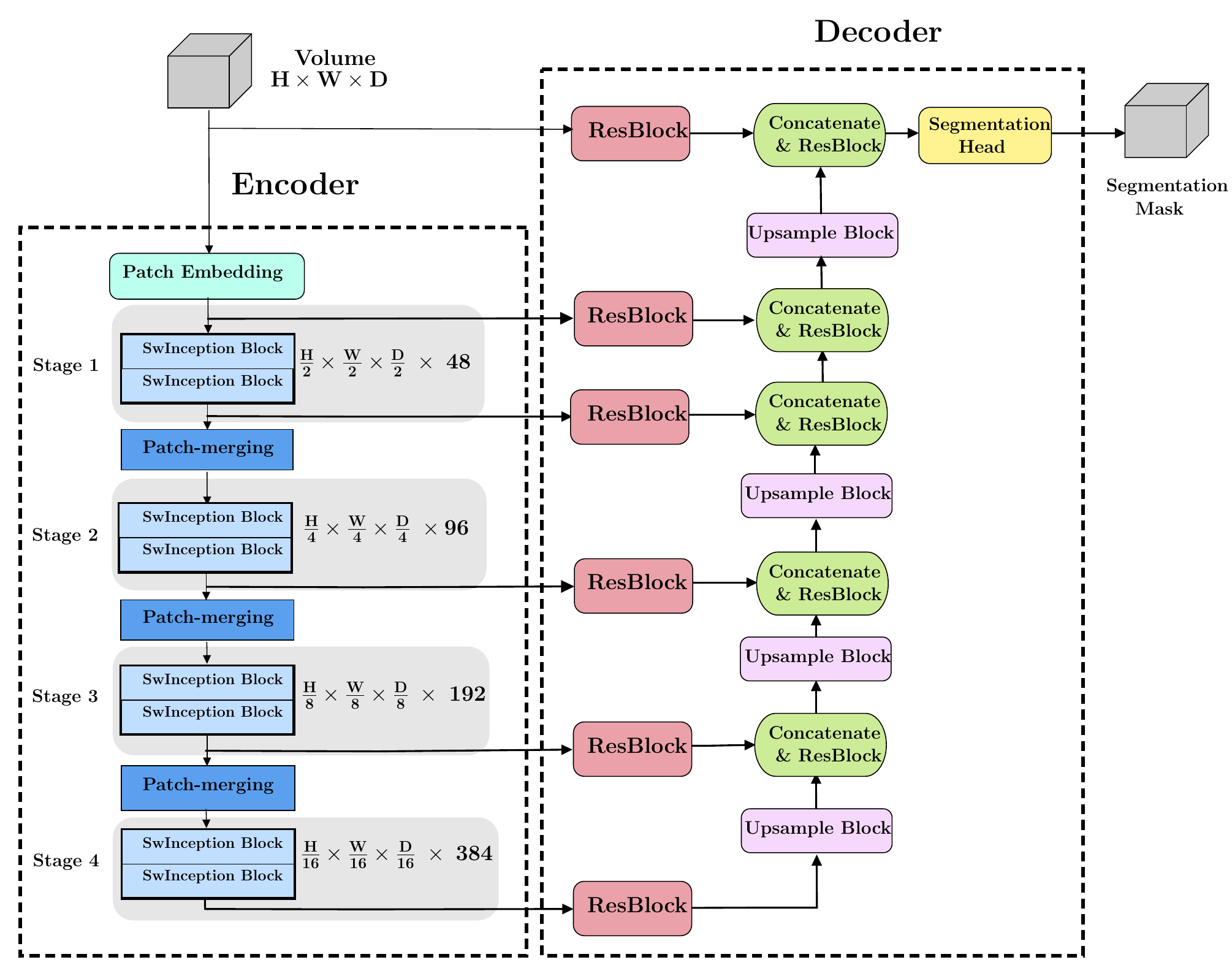}
\caption{An overview of the SwInception architecture for volumetric segmentation.}
\label{fig:swinception_arch}
\end{figure*}


\subsection{The SwInception encoder}
\label{sec:swinception_encoder}

As illustrated in Fig.~\ref{fig:swinception_arch}, the SwInception encoder consists of a patch embedding layer and four stages of SwInception blocks. The patch embedding layer is implemented as a convolutional layer comprising 48 filters, with both the stride and patch size set to two and no activation function. Each stage consists of two sequential SwInception blocks and maintains the same resolution throughout the entire stage. Between each stage, a patch merging block is used to downsample the resolution with a factor of 8 while doubling the output dimension. The patch merging block differs from the original Swin by utilizing a convolutional layer with overlapping filters as proposed in \cite{zhou_nnformer_2022}, which allows for stronger representations at the subsequent stages.

\subsubsection{The SwInception block. }
\label{sec:swinception_block}

\begin{figure}[ht]
\centering
    \begin{subfigure}{0.5\textwidth}
        \centering
        \includegraphics[width=0.6\linewidth]{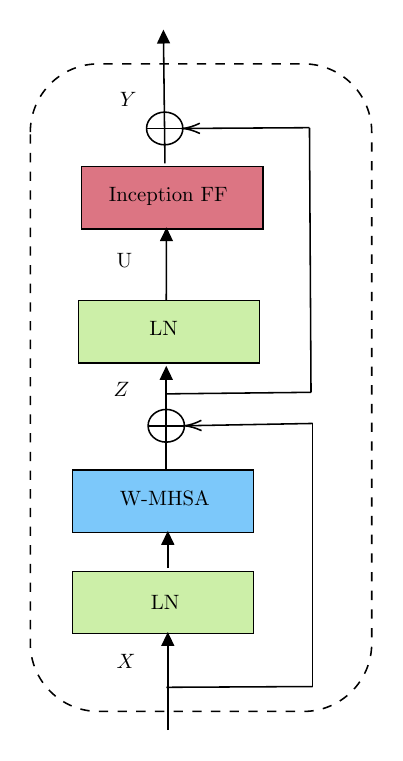}
        \caption{ }
        \label{fig:swinception_block}
    \end{subfigure}%
    \begin{subfigure}{0.5\textwidth}
        \includegraphics[width=0.9\linewidth]{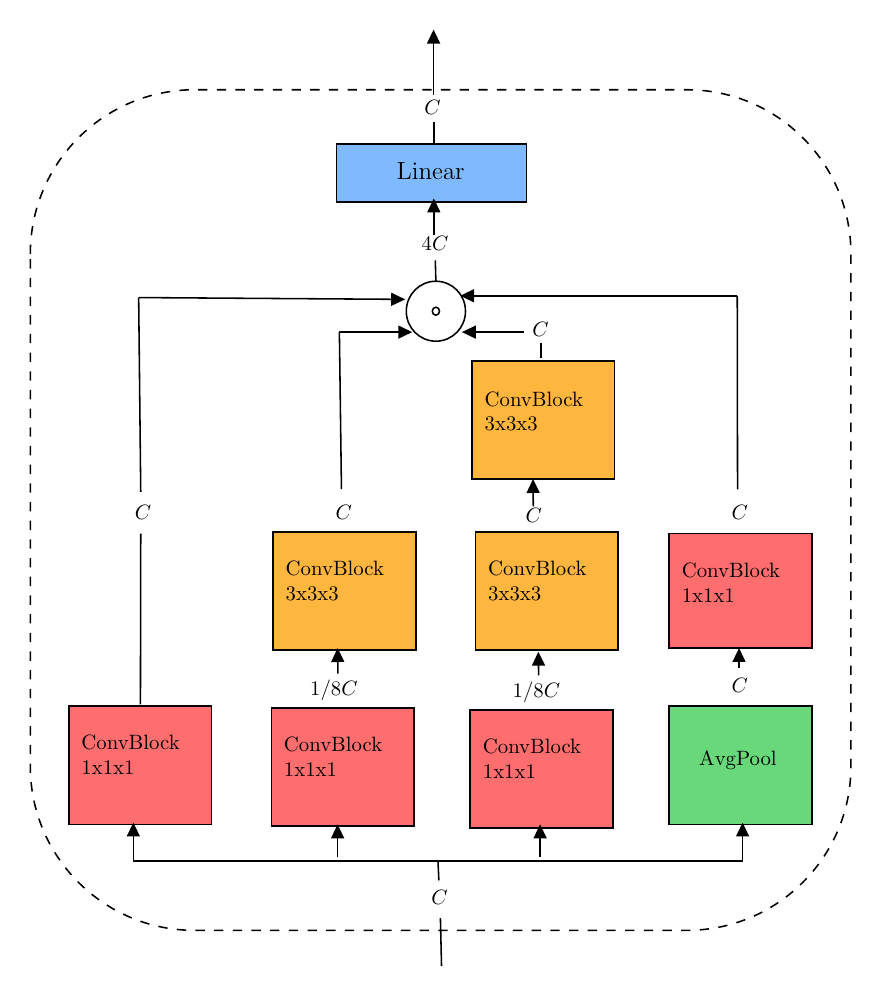}   
        \caption{ }
        \label{fig:inception}
    \end{subfigure}

\caption{(a) A SwInception block with layer normalization (LN), windowed multi-head self-attention (W-MHSA), and the Inception feed-forward block. (b) The Inception feed-forward block. Here $C$ denotes the number of channels.}
\label{fig:swinception} 
\end{figure}

SwInception is the first model that incorporates convolutions directly into the window multi-head self-attention (W-MHSA) mechanism from the Swin transformer~\cite{swin}. The convolutions are introduced in the feed-forward layer of the transformer block and can therefore improve the features used in later attention operations. More specifically, SwInception uses Inception \cite{szegedy_going_2014,szegedy_rethinking_2015} blocks as illustrated in Figs.~\ref{fig:swinception}~and~\ref{fig:inception}. 

Incorporating the Inception block in the Swin transformer introduces several advantageous features. First, convolutions enforce a stronger inductive bias toward locality in the architecture, reducing convergence time and lowering data requirements, which is particularly relevant for transformers. Second, convolutions increase the receptive field of the Swin transformer by expanding the number of tokens that communicate in each block. Third, convolutions in a transformer-based architecture can provide a stronger positional encoding and improve translation-equivariance, as shown in \cite{chu_conditional_2021}.
Incorporating the Inception block into the Swin transformer architecture results in enhanced performance and efficiency, albeit with a slight increase in parameter count. This is alleviated using bottlenecks at the beginning of the more expensive Inception branches. 

\subsubsection*{}
The Inception block serves as an intermediate solution between basic convolutional layers and depth-wise convolutional layers. While depth-wise convolutions are computationally inexpensive and less prone to overfitting, they cannot capture features that require input from multiple channels. In contrast, regular convolutions offer high capacity but can be expensive to use in transformer architectures. The Inception block strikes a balance by incorporating branches that operate on a subset of input channels, similar to depth-wise convolutions while retaining the ability to capture multi-channel features. By incorporating kernels of different sizes in the branches, our model can utilize features at varying scales within each transformer block, which proves advantageous for tasks like tumor detection involving objects of diverse sizes. Furthermore, both the pooling and convolution branches have local invariance to translations, enhancing this property of sparse vision transformers. It should be noted that in our research, we explored many different variations of Inception branches, convolutions, and depth-wise convolutions, but we are presenting the one we deemed best.

\subsubsection{SwInception block implementation. }
The output of SW-MHSA is reshaped into a volume and processed in parallel by each of the four Inception branches employed by SwInception. Let us recall that the MLP used in the feed-forward layer of Swin and most other transformer architectures utilize two sequential linear layers with an inverted bottleneck ratio of 4. From Fig.~\ref{fig:inception}, we observe that if the number of filters in the 1x1x1 branch is set to 4C and the number of filters in every other branch is zero (meaning that those branches are not used), we obtain the regular Swin MLP. By adjusting the number of filters in each branch, we can adjust how much relative weight to give each branch and thus how similar the SwInception block is to an MLP. In our research, we explored several types of weightings and found that equal branch weightings gave the strongest performance. 

Each Inception branch consists of convolutional blocks that include a convolutional layer with stride 1 and no dilation, batch normalization layer, and a GELU activation layer. The 1x1x1 branch comprises a single 1x1x1 convolutional block. The 3x3x3 branch first bottlenecks the feature channels by $\frac{1}{8}$ via a 1x1x1 convolutional block before applying a 3x3x3 convolutional block. The 5x5x5 branch has the same bottleneck as the 3x3x3 branch, followed by two 3x3x3 convolutional blocks. The pooling branch incorporates a 3x3x3 average pooling layer followed by a 1x1x1 convolutional block. The output of each branch is concatenated and reshaped into tokens before a final linear layer reduces the dimensionality back to the original size. 

The output $\boldsymbol{Y}$ of a SwInception block can be written as follows. Let $\boldsymbol{X}$ be the input to a SwInception block, $\text{LN}$ the layer normalization, and $\text{B}_i$, the $i$th Inception branch with $i \in \{1,2,3,4\}$. The shifted window multi-head self-attention is denoted by $\text{SW-MHSA}$ and shifts the windows every other block. Then, according to Figs.~\ref{fig:swinception}~and~\ref{fig:inception},
\begin{equation}
\begin{aligned}
    & \boldsymbol{Z} = \text{SW-MHSA}(\text{LN}(\boldsymbol{X})) + \boldsymbol{X}, \\
\end{aligned}
\end{equation}
\begin{equation}
\begin{aligned}
    & \boldsymbol{U} = \text{LN}(Z), \\
\end{aligned}
\end{equation}
\begin{equation}
\begin{aligned}
    & \boldsymbol{V} = \text{Concat}(\text{B}_1(\boldsymbol{U}), \text{B}_2(\boldsymbol{U}), \text{B}_3(\boldsymbol{U}), \text{B}_4(\boldsymbol{U})), \\
\end{aligned}
\end{equation}
\begin{equation}
\begin{aligned}
    & \boldsymbol{Y} = \text{Linear}(\boldsymbol{V}) + \boldsymbol{Z}. 
\end{aligned}
\end{equation}

\subsection{The SwInception decoder}
\label{sec:swinception_decoder}

The SwInception decoder is based on the decoder used in SwinUNETR \cite{tang_self-supervised_2022} but introduces a few important modifications. A complete diagram of the decoder model can be found in Fig.~\ref{fig:swinception_arch}. The decoder produces a segmentation map by utilizing the multi-scale features obtained from the encoder through the lateral skip connections. Each feature is first fed through a residual block using two 3x3x3 convolutional layers and concatenated with the features from the layer below. The resulting volume is upscaled with an upsampling block and then sent through a second residual block. This process is repeated for every layer $l_i, i \in \{0, 1, 2, 3, 4\}$ where $l_0$ is the patch embedding layer. The residual convolutional block comprises two 3x3x3 convolutional layers and a residual connection. The upsampling block utilizes a single 2x2x2 transposed convolutional layer. Instance normalization and PReLU activation are also employed in both block types.

What differentiates the SwInception decoder from the SwinUNETR decoder is that the extracted features are taken prior to the patch-merging step. This yields higher resolution features, obviates one upsampling step in the decoder, and eliminates the patch-merging step after the final SwInception block. Due to this reduction in the number of patch-merging operations, we also utilize a more expensive but efficient convolutional patch-merging strategy as mentioned in Section~\ref{sec:swinception_encoder}. The decoder still works on multi-scale features but at a higher resolution, which is beneficial for segmenting small objects. Additionally, the extracted feature vectors from the encoder are lower-dimensional, resulting in fewer decoder parameters as the convolutional filters become smaller.

\section{Experiments and Results}
\label{sec:experiments}
For an extensive evaluation of the capabilities of the SwInception architecture for medical volumetric segmentation, we utilize the Beyond the cranial vault (BTCV) \cite{gibson_automatic_2018} dataset as well as all ten datasets in the Medical Segmentation Decathlon challenge (MSD) \cite{antonelli_medical_2022}.

We present results from two SwInception versions. One uses the hyperparameters, pre-trained weights, and code from SwinUNETR \cite{tang_self-supervised_2022}, and we refer to that paper for details. As SwInception has several additional layers not included in the SwinUNETR pre-trained weights, they have been loaded in a non-strict fashion and frozen for the first 25 epochs.
The other version, denoted by SwInception*, is an optimized SwInception model that utilizes pre-trained SwInception weights and hyperparameters optimized specifically for SwInception. The pre-training was performed using the same methodology and datasets as presented in the SwinUNETR paper but using a SwInception encoder. Detailed hyperparameters can be found in Appendix~1.2 in the supplementary material.

We compare our work to the top three ranked models in the MSD challenge, SwinUNETR \cite{tang_self-supervised_2022}, nnUNet \cite{isensee_nnu-net_2021} and DiNTS \cite{He_2021_CVPR}. We also perform a comparison using the number one ranked solution, the Universal Model~\cite{liu2023clip}, with both SwInception and SwinUNETR as the base segmentation model.
All models used for comparison have been trained using their respective shared code, weights, and optimal hyperparameters. The parameter counts for the respective networks are as follows: nnUNet has 30M, SwinUNETR and SwInception each contain 63M, and DiNTS holds 152M.

The comparisons of architectures for medical volumetric segmentation in Sections~\ref{sec:btcv}~and~\ref{sec:msd} use averaged 5-fold cross-validation while the ablation studies are done on single folds. We have not compared performance on test data for two main reasons. First, the challenges is now closed and does not accept new submissions. Second, SwinUNETR \cite{tang_self-supervised_2022,swinunetr_private} and other models use extensive post-processing. The details regarding their post-processing steps are often unknown and, therefore, not reproducible. Considering that we want to compare model performance and not post-processing performance, we have opted for cross-validation as our method of choice.

All models are implemented in Python using the open-source libraries PyTorch and MONAI. Models have been trained with 4 A100 GPUs on a single node using mixed precision.




\subsection{Medical Segmentation Decathlon}
\label{sec:msd}

The MSD challenge \cite{antonelli_medical_2022} contains 6 CT datasets and 4 MRI datasets. Each dataset/task has its own training and test data, and the challenge covers a wide range of segmentation tasks for organs and lesions. The number of samples in the training sets ranges between 20 volumes (Heart) to 484 volumes (Brain).

We compare SwInception, SwInception*, and the top three models in the challenge: SwinUNETR, nnUNet, and DiNTS. For each task, we evaluate the models through 5-fold cross-validation over all training data. The averaged results across all folds can be found in Table~\ref{tab:msd}; for detailed results and the specific hyperparameters used by SwInception* we refer to the supplementary material. For MRI tasks, no pre-trained weights are used for any model. No post-processing has been used for any of the listed models.

The results show that even the baseline SwInception model outperforms the previous state-of-the-art models when looking at average performance over all tasks. In particular, a significant increase can be observed both at MRI tasks such as prostate segmentation and CT tasks like lung cancer.

The optimized SwInception* increases the gap further with large increases at several tasks.
Using SwInception's pre-trained weights grants major improvements on CT tasks, possibly due to the transformer block weights being properly optimized to leverage the locally enhanced features. The difference is particularly clear on cancer segmentation tasks such as colon, liver, and pancreas, which are some of the most difficult segmentation tasks in the challenge.

\begin{table*}[ht]
\setlength{\tabcolsep}{3pt}
\centering
\caption{Cross-validation performance on MSD from SwInception, SwinUNETR, nnUNet, and DiNTS.}\label{tab:msd}
\begin{tabular*}{0.85\textwidth}{c|c|c|c|c|c}
\multicolumn{6}{c}{} \\
Task & Brain Tumour & Heart &  Liver & Hippocampus & Prostate \\
\hline
Metric & \multicolumn{5}{c}{Mean Dice}\\
\hline
DiNTS & 72.63 & 92.20 & 72.21 & 88.13 & 70.60 \\
nnUNet & 74.03 & \textbf{93.30} & 76.84 & 89.04 & 71.72 \\
SwinUNETR & 74.26 & 90.78 & 78.69 & 87.08 & 71.59 \\
SwInception & 74.49 & 92.57 & 79.22 & 87.34 & 73.01 \\
SwInception* & \textbf{74.57} & 92.60 & \textbf{82.19} & \textbf{89.06} & \textbf{74.77} \\
\hline
\multicolumn{6}{c}{} \\
\end{tabular*}
\centering
\begin{tabular*}{0.85\textwidth}{c|c|c|c|c|c||c}
Task & Lung & Pancreas & Hepatic Vessel & Spleen & Colon & All\\
\hline
Metric & \multicolumn{6}{c}{Mean Dice}\\
\hline
DiNTS & 60.35 & 57.98 & 59.94 & 94.68 & 37.54 & 70.63 \\
nnUNet & 64.09 & 66.58 & \textbf{66.58} & 95.35 & 41.53 & 73.91 \\
SwinUNETR & 64.68 & 62.97 & 62.72 & 95.66 & 42.74 & 73.12 \\
SwInception & 66.73 & 64.57 & 64.10 & 96.24 & 43.73 & 74.20 \\
SwInception* & \textbf{68.03} & \textbf{67.03} & 66.33 & \textbf{96.39} & \textbf{48.19} & \textbf{75.92}\\
\hline
\end{tabular*}
\end{table*}

\subsection{Beyond the Cranial Vault}
\label{sec:btcv}

BTCV \cite{gibson_automatic_2018} is an abdomen multi-organ segmentation dataset first released in conjunction with MICCAI 2015 comprised of 30 volumes, and the data is collected from patients with either colorectal cancer or ventral hernia.

We perform a 5-fold cross-validation comparison between SwInception, SwInception*, SwinUNETR, nnUNet, and DiNTS. The results can be found in Table~\ref{tab:btcv}. For detailed per-organ segmentation results, we refer to Appendix~1.1 in the supplementary material. 

The experiments show that SwInception* outperforms all other models on average. The nnUNet architecture also shows great performance, indicating the importance of larger crops for multi-organ segmentation in large volumes. The experiments suffer from a large variance with big differences in performance between different folds due to the small size of the dataset.

\begin{table*}[ht]
\setlength{\tabcolsep}{3pt}
\centering
\caption{BTCV results from SwInception and SwinUNETR.}\label{tab:btcv}
\begin{tabular}{c|c|c|c|c|c|c}
\multicolumn{7}{c}{} \\
 \multicolumn{1}{c|}{Fold} & \multicolumn{1}{c|}{1} & \multicolumn{1}{c|}{2} & \multicolumn{1}{c|}{3} & \multicolumn{1}{c|}{4} & \multicolumn{1}{c||}{5} & \multicolumn{1}{c}{All} \\
\hline
\multicolumn{1}{c|}{Metric} & \multicolumn{6}{c}{Mean Dice} \\
\hline
\multicolumn{1}{c|}{DiNTS} & 77.11 & 72.49 & 76.57 & 75.78 & 71.04 & 74.60 \\
\multicolumn{1}{c|}{nnUNet} & 82.91 & \textbf{79.33} & 81.32 & 82.15 & 73.57 & 79.86 \\
\multicolumn{1}{c|}{SwinUNETR} & 80.64 & 71.78 & 79.19 & 78.01 & 77.75 & 77.14 \\
\multicolumn{1}{c|}{SwInception} & 82.53 & 71.61 & 80.49 & 80.06 & \textbf{78.67} & 78.67  \\
\multicolumn{1}{c|}{SwInception*} & \textbf{84.15} & 73.00 & \textbf{82.45} & \textbf{83.14} & 77.82 & \textbf{80.11} \\
\hline
\end{tabular}
\end{table*}

\subsection{Ablation study on encoder and decoder combinations}
\label{sec:ablation}

We investigate the effect the separate parts of the SwInception architecture have on performance by comparing different combinations of the encoder, decoder, and encoder patch-merging strategy. The experiments are all performed on the Decathlon Prostate dataset, which is a challenging task with very large inter-subject variability. All models have been trained from scratch without pre-trained weights. For a fair comparison between the Swin and SwInception encoders, we also include experiments where the inverted bottleneck ratio for the feed-forward layer in Swin has been increased from 4.0 to 7.0 such that the number of parameters is roughly equal for both models. 

From the results in Table~\ref{tab:ablation}, it can be observed that changing the encoder from Swin to SwInception always increases performance, even when compared to a Swin model with the same number of parameters in the feed-forward layer. The experiments also show that the proposed decoder improves performance for all encoder types with the added benefit of a lowered parameter count. The convolutional patch-merging strategy, described in Section~\ref{sec:swinception_encoder}, generally performs better but at the cost of more parameters, specifically when using the SwinUNETR decoder, due to the final patch-merging operation being performed on feature maps with many channels.

\begin{table*}[ht]
\setlength{\tabcolsep}{3pt}
\centering
\caption{Ablation study over encoder, decoder, and patch-merging strategies on MSD Prostate.}\label{tab:ablation}
\begin{tabular*}{0.95\textwidth}{c|c|c|c|c||c}
\multicolumn{6}{c}{} \\
Encoder & Decoder & Patch-Merging & MLP-ratio & Params & Mean Dice  \\
\hline
Swin & SwinUNETR & Linear & 4.0 & 62.8M & 72.17 \\
Swin & SwinUNETR & Linear & 7.0 & 65.2M & 72.36 \\
SwInception & SwinUNETR & Linear & 4.0 & 64.9M & \textbf{72.81} \\
\hline
Swin & SwinUNETR & Conv & 7.0 & 72.6M & 73.48 \\
SwInception & SwinUNETR & Conv & 4.0 & 72.3M & \textbf{74.97} \\
\hline
Swin & SwInception & Linear & 4.0 & 59.2M & 73.99 \\
Swin & SwInception & Linear & 7.0 & 61.6M & 72.38 \\
SwInception & SwInception & Linear & 4.0 & 61.3M & \textbf{75.20} \\
\hline
Swin & SwInception & Conv & 7.0 & 63.4M & 72.99 \\
SwInception & SwInception & Conv & 4.0 & 63.1M & \textbf{75.33} \\
\hline
\end{tabular*}
\end{table*}

\subsection{The choice of Inception block}

We investigate the difference between adding an Inception block and depth-wise convolutional blocks to a Swin encoder. The architecture denoted as SwinDepth has two blocks of depth-wise convolutions, batch norms, and GELU activations in between the two linear layers in the Swin feed-forward layer, similar to \cite{xie_segformer_2021,guo_cmt_2022,yuan_incorporating_2021}. All models were trained on single folds from three challenging MSD datasets with the SwinUNETR decoder and the linear patch merging strategy. All models were trained from scratch without pre-trained weights. The inverted bottleneck ratio in the MLP was increased to 7.0 for Swin and SwinDepth to make the parameter count equivalent for all models.

The results in Table~\ref{tab:ablation_inception block} show that introducing depth-wise convolutions can improve performance for the baseline Swin encoder on specific datasets. However, the improvements for the SwInception encoder are both larger and more consistent while introducing only a minor increase in parameter count.

\begin{table}[ht]
\setlength{\tabcolsep}{3pt}
\centering
\caption{A comparison between Swin encoders with feed-forward layers using Inception and Depth-Wise convolutions.}\label{tab:ablation_inception block}
\begin{tabular*}{0.5\textwidth}{c|c|c|c}
\multicolumn{4}{c}{} \\
Task & Liver & Pancreas & Colon \\
\hline
Metric & \multicolumn{3}{c}{Mean Dice} \\
\hline
Swin & 83.84 & 76.02 & 73.55 \\
SwinDepth & 84.25 & 75.78 & 76.14  \\
SwInception & \textbf{84.84} & \textbf{77.40} & \textbf{77.08} \\
\hline
\end{tabular*}
\end{table}

\subsection{Visual comparison}

To further analyze the differences between SwInception* and the models evaluated on MSD, we present a visual comparison in Table~\ref{tab:visual_comp}. We can see that while organ segmentation results are very similar regardless of the model, SwInception improves the rate of true positives for cancer segmentation without adding any false positives. In general, the segmentations from SwInceptions are smoother and less fragmented, which improves the segmentation accuracy and reduces the need for post-processing.

\begin{table*}[ht]
\setlength{\tabcolsep}{2pt}
    \centering
    \caption{Visual comparison using examples from MSD. A. Liver cancer. B. Colon cancer. C. Pancreatic cancer. Green denotes organ segmentation, and yellow denotes cancer segmentation for all examples.}
    \label{tab:visual_comp}
        \begin{tabular}{cccccc}
        & Ground Truth & SwInception* & SwinUNETR & nnUNet & DiNTS \\[-1pt]
             A & \includegraphics[width=0.17\linewidth]{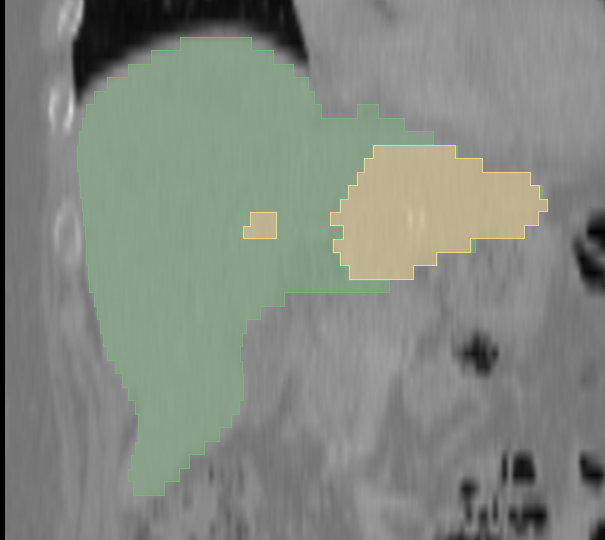}
            & \includegraphics[width=0.17\linewidth]{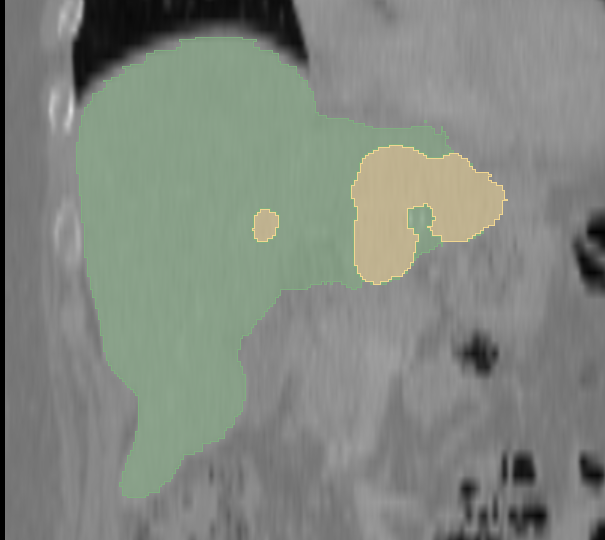}
            & \includegraphics[width=0.17\linewidth]{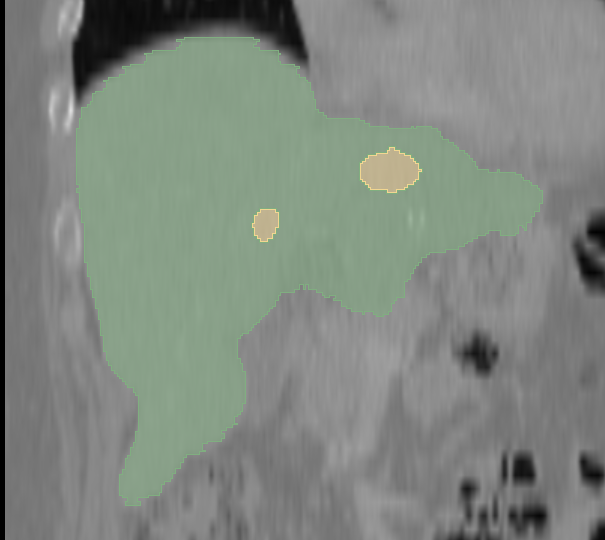}
            & \includegraphics[width=0.17\linewidth]{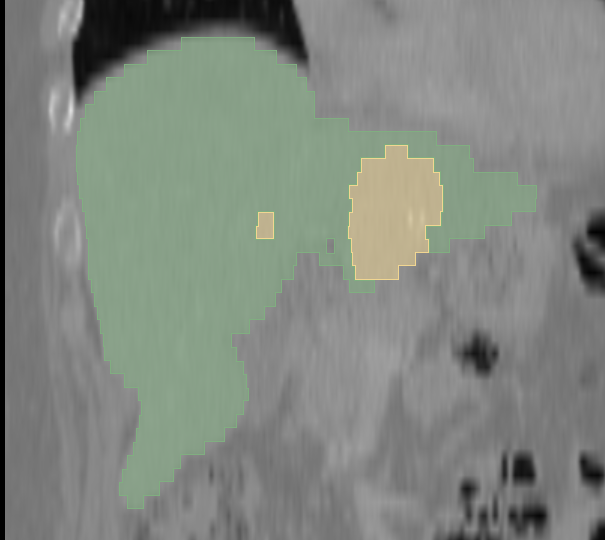}
            & \includegraphics[width=0.17\linewidth]{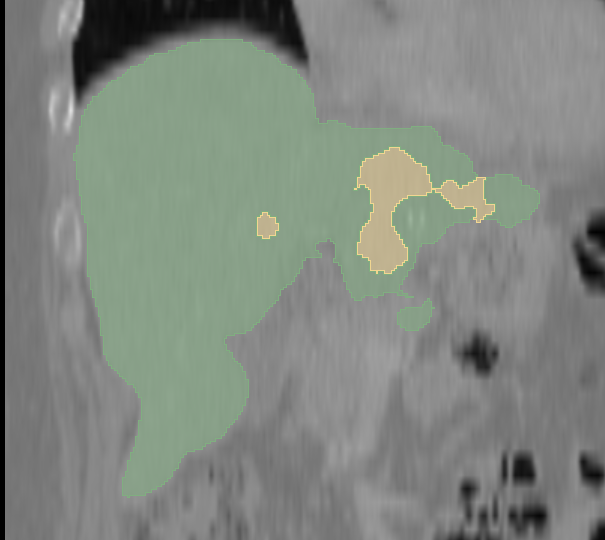}\\[-1pt]
            B & \includegraphics[width=0.17\linewidth]{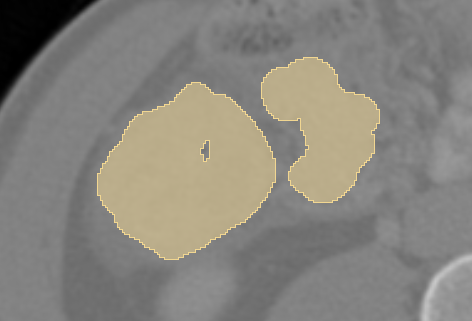}
            & \includegraphics[width=0.17\linewidth]{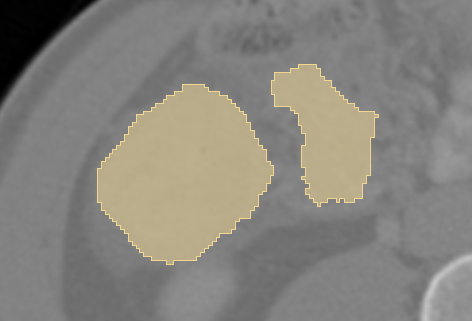}
            & \includegraphics[width=0.17\linewidth]{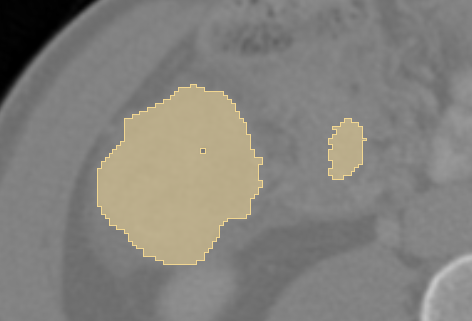}
            & \includegraphics[width=0.17\linewidth]{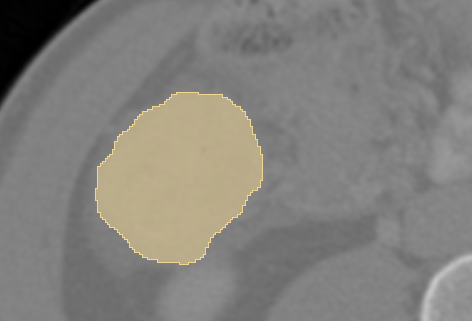}
            & \includegraphics[width=0.17\linewidth]{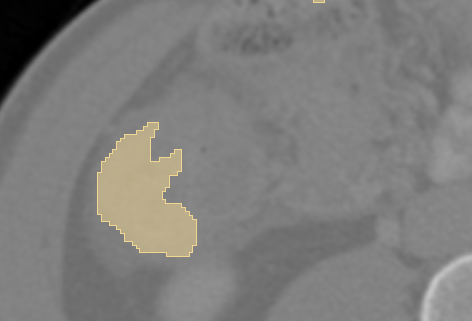}\\[-1pt]
            C & \includegraphics[width=0.17\linewidth]{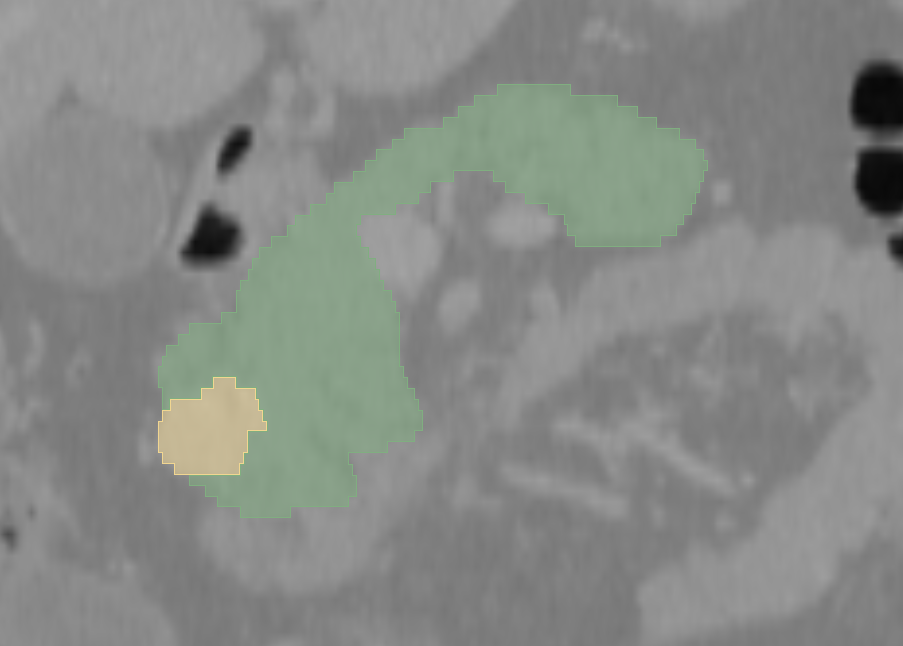}
            & \includegraphics[width=0.17\linewidth]{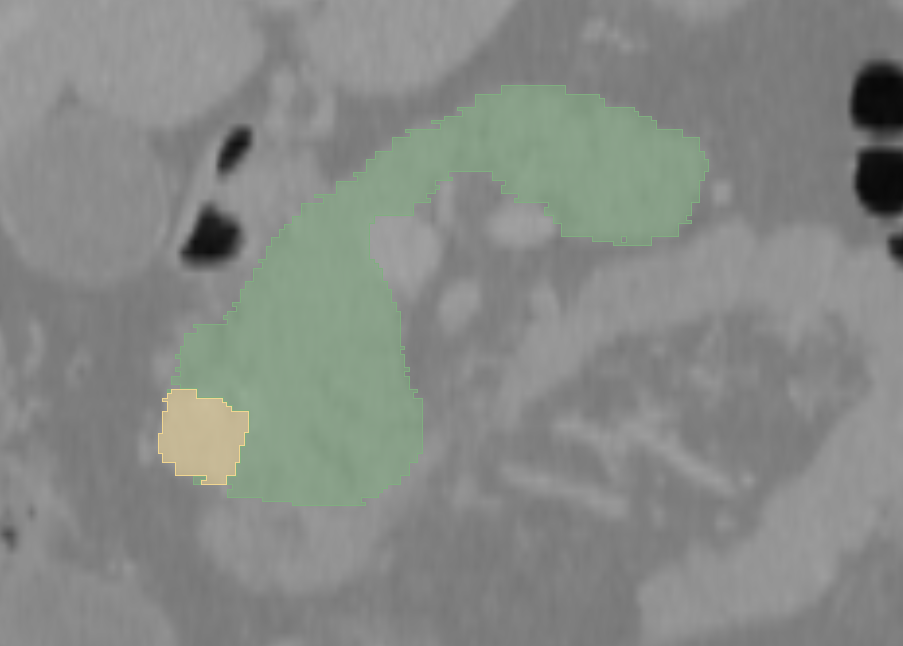}
            & \includegraphics[width=0.17\linewidth]{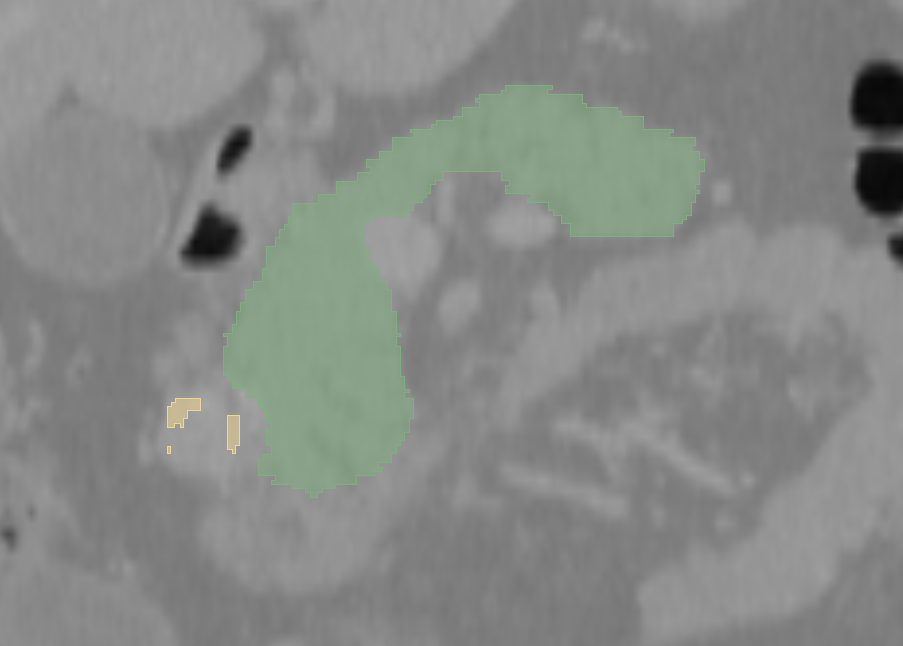}
            & \includegraphics[width=0.17\linewidth]{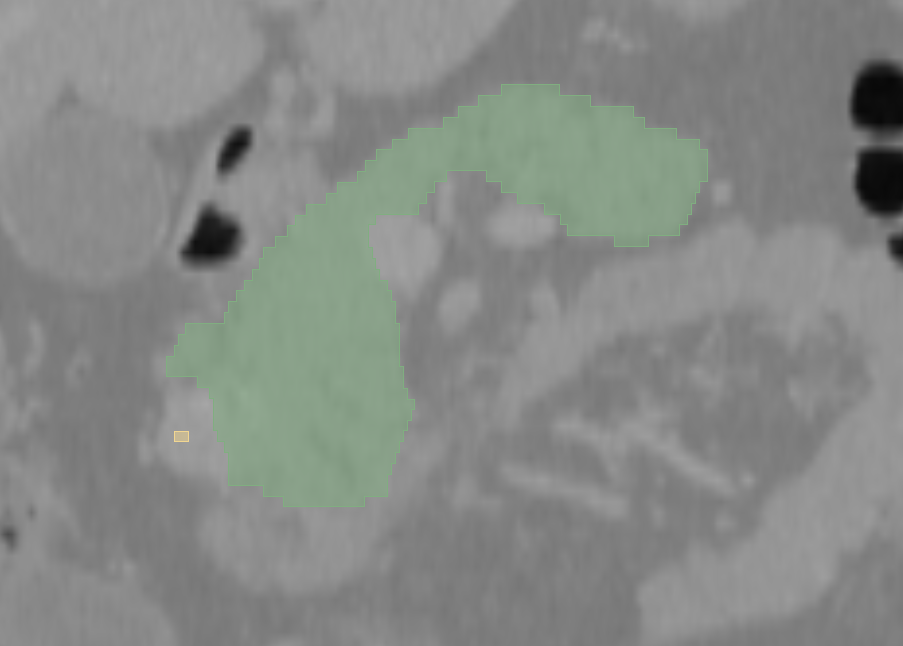}
            & \includegraphics[width=0.17\linewidth]{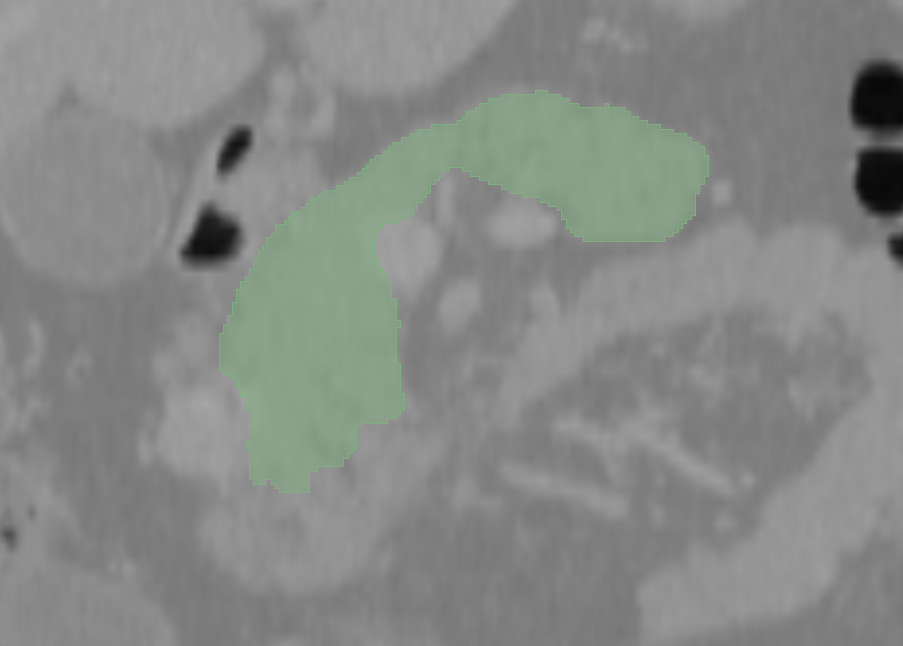}\\[-1pt]
        \end{tabular}%
\end{table*}

\subsection{Model performance on smaller datasets}
\label{sec:small_datasets}

In Table \ref{tab:small-dset}, a comparison between SwinUNETR and SwInception on several differently sized subsets of Decathlon training data can be found. All the data that was not used in the subset has been utilized as validation data. Both models have been trained as in Section~\ref{sec:msd} using SwinUNETR hyperparameters and no pre-trained weights. The results demonstrate that SwInception outperforms SwinUNETR on all subset sizes. Notably, the relative difference grows significantly for the smaller datasets, especially for the more difficult task of segmenting colon cancer. From an 11.8\% relative increase on the colon subset containing half of the dataset to a 34.5\% relative increase on the smallest colon subset with only 10\% of the data. As one effect of a stronger inductive bias is to reduce the data required to train models to convergence, these results align with the theorized advantages that the multi-branch convolutions bring to the Swin encoder.

\begin{table}[ht]
\centering
\setlength{\tabcolsep}{3pt}
\caption{A comparison between SwInception and SwinUNETR on small datasets.}\label{tab:small-dset}
\begin{tabular*}{0.7\textwidth}{c|c|c|c|c|c|c}
\multicolumn{7}{c}{} \\
\multicolumn{1}{c|}{Task} & \multicolumn{3}{c|}{Hippocampus} & \multicolumn{3}{|c}{Colon}\\
\hline
\multicolumn{1}{c|}{Subset size} & 50\% & 25\% & 10\% & 50\% & 25\% & 10\%\\
\hline
\multicolumn{1}{c|}{Metric} & \multicolumn{3}{|c|}{Mean Dice} & \multicolumn{3}{|c}{Tumor Dice}\\
\hline
\multicolumn{1}{c|}{SwinUNETR} & 91.09 & 90.67 & 90.15 & 38.64 & 27.05 & 15.18\\
\multicolumn{1}{c|}{SwInception} & \textbf{91.71} & \textbf{91.41} & \textbf{90.95} & \textbf{43.23} & \textbf{34.02} & \textbf{20.42}\\
\hline
\end{tabular*}
\end{table}

\subsection{SwInception as a backbone in UniversalModel}
\label{sec:universal_model}

The UniversalModel, utilizing upon the CLIP framework \cite{liu2023clip}, introduces a pre-training strategy that facilitates the integration of multiple diverse medical datasets in a supervised pre-training phase. Utilizing SwinUNETR as its backbone segmentation model, it currently stands as the SOTA method for the Medical Segmentation Decathlon. Table~\ref{tab:universalmodel} presents extensive cross-validation results across all MSD CT tasks and BTCV, employing SwinUNETR and SwInception backbones. The models were pre-trained following the methodology outlined in the paper, excluding any dataset containing MSD or BTCV data. The codebase and hyperparameters remain consistent with the original paper, with two modifications: post-processing was disabled, and learning rates were optimized for specific tasks due to the smaller volume of pre-training data.

Our findings consistently demonstrate that employing SwInception as the base segmentation model outperforms SwinUNETR across diverse tasks. This performance improvement is particularly evident in challenging tasks that require the segmentation of small cancerous regions, such as those found in the Lung and Colon. Some of the cancer tasks show worse performance compared to the results in Table~\ref{tab:msd}. This could be attributed to the fact that the pre-training data now contains very low number of cancer annotations after the exclusion of MSD data. Conversely, tasks such as Spleen and BTCV showcase substantial performance enhancements, aligning with the prevalence of general organ segmentation data in the majority of the pre-training datasets.

\begin{table*}[ht]
\setlength{\tabcolsep}{3pt}
\centering
\caption{UniversalModel with SwinUNETR and SwInception as backbones on Decathlon CT tasks and BTCV}\label{tab:universalmodel}
\begin{tabular*}{\textwidth}{c|c|c|c|c|c|c|c}
\multicolumn{8}{c}{} \\
Task & Liver & Lung & Pancreas & Hepatic Vessel & Spleen & Colon & BTCV \\
\hline
Metric & \multicolumn{7}{c}{Mean Dice}\\
\hline
UM-SwinUNETR & 76.13 & 57.07 & 62.55 & 62.49 & 96.63 & 43.01 & 82.60 \\
UM-SwInception & \textbf{77.40} & \textbf{60.06} & \textbf{64.39} & \textbf{63.73} & \textbf{96.83} & \textbf{46.12} & \textbf{82.93} \\
\hline
\end{tabular*}
\end{table*}

\section{Conclusion}

In this paper, we have investigated the effect a stronger local inductive bias can have on the Swin architecture for small and medium-sized datasets and how to efficiently utilize encoder features for volumetric semantic segmentation. These investigations have resulted in a backbone encoder and a segmentation decoder that we name SwInception. This hybrid transformer-convolution architecture outperforms the previous state-of-the-art methods on competitive medical image segmentation challenges. Finally, we observe significant increases in performance on tiny datasets, possibly due to the stronger inductive bias introduced by the convolutional branches. 

\subsubsection{Acknowledgements}
The computations were enabled by resources provided by the National Academic Infrastructure for Supercomputing in Sweden and the Swedish National Infrastructure for Computing at Chalmers Centre for Computational Science and Engineering partially funded by the Swedish Research Council through grant agreements no. 2022-06725 and no. 2018-05973.
The computations were also enabled by the Berzelius resource provided by the Knut and Alice Wallenberg Foundation at the National Supercomputer Centre.

\bibliographystyle{splncs04}
\bibliography{mainbibetal}
\end{document}